\documentclass{IOS-Book-Article}

\usepackage{mathptmx}

\usepackage{graphicx}
\usepackage{longtable}
\usepackage{url}
\usepackage{amsmath}

\def\hb{\hbox to 10.7 cm{}}

\begin{document}

\pagestyle{headings}
\def\thepage{}

\begin{frontmatter}              

\title{Towards a Universal Neural Network Encoder for Time Series}


\author[A]{Joan Serr\`a%
\thanks{Corresponding author: Joan Serr\`a, Telef\'onica Research, Pl.~Ernest Lluch i Mart\'in 5, 08019 Barcelona; E-mail:
\url{joan.serra@telefonica.com}.}},
\author[B]{Santiago Pascual}
and
\author[A]{Alexandros Karatzoglou}

\address[A]{Telef\'onica Research, Barcelona}
\address[B]{Universitat Polit\`ecnica de Catalunya, Barcelona}

\begin{abstract}
We study the use of a time series encoder to learn representations that are useful on data set types with which it has not been trained on. The encoder is formed of a convolutional neural network whose temporal output is summarized by a convolutional attention mechanism. This way, we obtain a compact, fixed-length representation from longer, variable-length time series. We evaluate the performance of the proposed approach on a well-known time series classification benchmark, considering full adaptation, partial adaptation, and no adaptation of the encoder to the new data type. Results show that such strategies are competitive with the state-of-the-art, often outperforming conceptually-matching approaches. Besides accuracy scores, the facility of adaptation and the efficiency of pre-trained encoders make them an appealing option for the processing of scarcely- or non-labeled time series.
\end{abstract}

\end{frontmatter}


\section{Introduction}

Time series data present a number of characteristics that motivate specific processing strategies. Besides the importance of attribute ordering, temporal correlations, periodicities, and drifts~\cite{Han05BOOK}, time series algorithms typically deal with variable lengths, high-dimensional inputs, and scarcely labeled data. For instance, the UEA/UCR time series classification repository~\cite{Bagnall17WEB} contains data sets of sizes ranging between 40 and 16,637~instances, and lengths/dimensionalities between 24 and 2,709~samples. Apart from classification~\cite{Bagnall17DMKD}, other important tasks in time series are clustering~\cite{Liao05PR}, segmentation~\cite{Keogh04BOOKCHAP}, motif discovery~\cite{Mueen14WIRES}, anomaly detection~\cite{Chandola09ACS}, and forecasting~\cite{Hyndman13BOOK}.

In this paper, we study the use of an encoder to tackle the aforementioned challenges of variable length, high-dimension, and few labeled data. To overcome the first two challenges, we couple a time-wise attention mechanism with convolutional neural networks. The attention mechanism summarizes variable-length representations into fixed-length vectors, while convolutions deal with local/temporal correlations. To overcome the latter challenge, we propose to learn a universal network, trained with a variety of data sets, that can deal with new data types without further intervention or training. Overall, our objective is to develop and train an encoder network that converts variable-length time series to a fixed-length, low-dimensional representation which, when interchanged with the raw time series or other features extracted from it, improve a reference task. Importantly, we want the learned representations to generalize to unseen data types, with minimal or even no adaptation of the encoder network to the novel data. This last point, the generalization of learned representations to unseen data types, is an active area of research within machine learning which, to the best of our knowledge, has not received much attention in the time series domain. 

Although the usage of the proposed encoder and its representations aim at general time series problems, in this paper, we restrict ourselves to the problem of time series classification~\cite{Bagnall17DMKD}, as it allows for a clear and objective evaluation, and also well represents the aforementioned challenges in time series processing~\cite{Han05BOOK}. Moreover, there exist a reasonable amount and variety of time series classification data sets~\cite{Bagnall17WEB}, organized by data type, with which we can conveniently train an encoder and then test it with an unseen type. Under this setting, a pre-trained universal encoder should produce representations that are useful to automatically label, for instance, an electrocardiogram (ECG) data set, without having seen any ECG instance in the training phase. Such labeling should be performed with minimal adaptation to the target data or, in the extreme case, without any learning over such data.


\section{Related Work}

Multi-task learning~\cite{Caruana97ML}, in which commonalities across multiple related tasks are exploited to better solve some target task(s), has a long tradition. In the main setting, multi-task learning uses a shared representation that is learnt in parallel across several tasks, including the target one(s). This can be an unrealistic scenario, as target data sets may not be available beforehand, they may not have labels, or it simply may become unfeasible to re-train in parallel with all data sets every time we find a new target task~\cite{Thrun95RAS}. Transfer learning~\cite{Pratt93NIPS} is an interesting alternative, in which a pre-trained model is adapted to a new target task, with less effort and better results than training from scratch on the new task. Transfer learning typically does not reuse previous data in the adaptation step but, nonetheless, it assumes labeled data for the target task. Notice also that, under a sequential or lifelong learning scenario~\cite{Thrun95RAS}, repeated transfer learning may yield to the phenomenon of catastrophic forgetting~\cite{McCloskey89PLM}, in which the knowledge of previous tasks progressively vanishes.

In order to have sufficient knowledge to accomplish any task, and in order to be applicable in the absence of labeled data or even without adaptation/re-training, researchers have been increasingly adopting the generic concept of universal encoders, specially within the text processing domain~\cite{Cer18ARXIV,Conneau17EMNLP,Ha16IWSLT} (note that related concepts also exist in other domains~\cite{Chung16IS,Finn17ICML,Oord17ARXIV}). The basic idea is to train a model (the encoder) that learns a common representation which is useful for a variety of tasks and that, at the same time, can be reused for novel tasks with minimal or no adaptation. While it would seem that classical autoencoders and other unsupervised models should perfectly fit this purpose, recent research in sentence encoding shows that, with current means, encoders learnt with a sufficiently large set of supervised tasks~\cite{Conneau17EMNLP}, or mixing supervised and unsupervised data~\cite{Cer18ARXIV}, consistently outperform their purely unsupervised counterparts. 

Despite time series classification offers an interesting testbed for universal encoders, to the best of our knowledge, only Malhorta et al.~\cite{Malhorta17ESANN} learn time series encoders whose outputs are later exploited to perform new classification tasks. In particular, they consider seq2seq~\cite{Sutskever14NIPS} autoencoders, and train them to reconstruct time series, either with single or multiple data sets. Adaptation to the new (supervised) data set is done through support vector machine classifiers with radial basis function kernels. They report accuracies marginally over the typical nearest-neighbor classifier using a dynamic time warping (DTW) distance. 

Deep neural networks are progressively being introduced to the problem of time series classification, with promising results~\cite{Cui16ARXIV,Hatami17ARXIV,Wang16ARXIV}. However, due to the diversity in time series lengths and the low number of instances in the training sets (often under 100), these type of algorithms seem to struggle to catch up with more competitive approaches~\cite{Lines16ICDM}. In general, ensemble approaches with multiple classifiers, features, and distances are the most competitive ones~\cite{Bagnall17DMKD}. Two successful algorithms of this kind are COTE~\cite{Bagnall15TKDE} and HIVE-COTE~\cite{Lines16ICDM}. Canonical baseline approaches using the raw time series are based on nearest-neighbor classifiers with elastic distances~\cite{Serra14KBS}, such as the aforementioned DTW distance, or feature-based classifiers on top of the raw time series~\cite{Bagnall17DMKD}. Their accuracies are always significantly below the ones achieved by competitive ensemble methods.


\section{Towards a Universal Encoder for Time Series}

\subsection{Architecture}

In the design of the encoder network we strive for simplicity and efficiency. That is, we strive for a model that is both conceptually straightforward and computationally lightweight. The former is interesting for implicit regularization and ease of explanation, understanding, and deployment. The latter is important for pre-trained model transfer between users and speed of operation.

\begin{figure}[t]
	\includegraphics[width=1\linewidth]{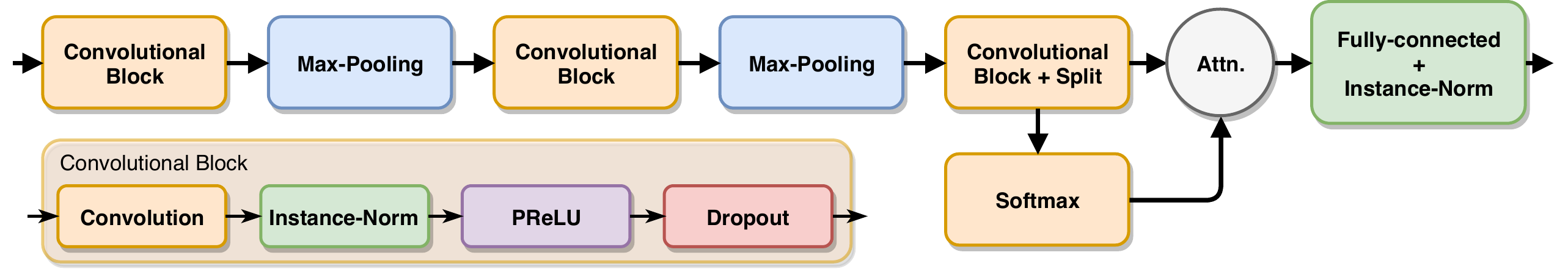}
	\caption{Architecture diagram of the proposed encoder (top) and the convolutional block (bottom left).}
	\label{fig:diagram}
\end{figure}

The model we consider as encoder is a standard convolutional network, with a convolutional attention mechanism to summarize the time axis, and a final fully-connected layer to set the desired representation dimensionality (Fig.~\ref{fig:diagram}). The convolutional network is formed by three convolutional blocks with two 2-factor max-pooling layers between them. A convolutional block is formed by a 1-dimensional convolution, followed by an instance normalization layer~\cite{Ulyanov16ARXIV}, a parametric rectified linear unit (PReLU)~\cite{He15ICCV} activation, and a dropout layer (Fig.~\ref{fig:diagram}, bottom left). After the first part of the network, half of the filters are input to a time-wise softmax activation, which acts as an attention mechanism for the other half of the filters. That is, for a single filter, 
\begin{equation}
h = \textbf{h} \cdot \textbf{a},
\label{eq:attention}
\end{equation}
where $\cdot$ denotes a dot product, $\textbf{h}$ is the result of a single 1-dimensional convolutional filter over a time-wise signal, and $\textbf{a}$ is the time-wise attention vector (independent for each filter). The result of the attention mechanism for all filters is finally passed through a fully-connected and an instance normalization layers (Fig.~\ref{fig:diagram}, top right). We found instance normalization to facilitate training and to provide more consistent value ranges in the encoder's output. The dimensionality of the output is denoted by $k$, which is a parameter whose impact we study below (Sec.~\ref{sec:res_encoding_dim}).

\subsection{Implementation Details}
\label{sec:enc_details}

For the three convolutional layers we use\footnote{Unless stated otherwise, we use PyTorch~\cite{Paszke17NIPSW} version 0.3.1 with default parameters.}, respectively, 128, 256, and 512~filters, with kernels of 5, 11, and 21 (stride of 1), and same-length padding of 3, 5, and 10. Instance normalization includes the affine transformation~\cite{Ioffe15ICML}, and PReLUs are multi-parametric, that is, they have one slope parameter per filter. We use a dropout of 0.2 in all layers. Half of the 512~filters of the last convolutional layer are input to the softmax layer and later used to compute the filter-wise dot product with the remaining half (Eq.~\ref{eq:attention}). 

\subsection{Training}
\label{sec:enc_training}

For learning the weights of the encoder network, we take all training data sets one by one and backpropagate the error on each data set batch-wise. That is, for every data set, we take a single batch of examples and do a forward pass with an extra classification, fully-connected layer that learns to map the encoder representation to the number of classes of the given data set. We then take the softmax of that output and measure the error with categorical cross-entropy. This strategy corresponds to a so-called multi-head output~\cite{Bakker02JMLR}, typically used in multitask or sequential learning. After the forward pass, we backpropagate the error to both the extra layer and the encoder network. We repeat this process using single random batches for every single data set 20~times. That is, if there are $n$ data sets in the training set, we do $20n$ single-batch forward-backward passes. This defines a training epoch.

To update the weights of the networks we use plain stochastic gradient descent with a learning rate of 0.005. We reduce the learning rate by a factor of 3 if we do not observe an improvement in the validation loss for more than 10 epochs, and stop training when we hit a learning rate below $10^{-4}$. As validation loss we use an average of the per-data set losses, using all validation data. We use a batch size of 12.


\section{Evaluation Methodology}
\label{sec:eval}

\subsection{Data and Splitting}

To assess the quality of the representations, we consider the task of time series classification~\cite{Bagnall17DMKD}. In particular, we consider the 85~data sets of the UEA/UCR time series classification repository~\cite{Bagnall17WEB}. To assess the generalization capabilities of the learned representations, we form encoder train/test splits according to the data type. This way, at test time, we evaluate the encoder with a data type that has not been used for training. The repository contains 7~data types: electric devices (6 data sets), ECGs (7), image outlines (29), motion capture (14), sensor readings (16), spectrographs (7), and simulated/artificial data (6). Therefore, we follow a 7-fold training procedure. When learning the parameters of the encoder, we leave out all the data sets corresponding to one data type for testing, and split the rest of the data sets into train/validation following a per-data set, non-stratified 80/20\%~rule. At test time, we take the left out data sets (corresponding to one data type) and use the original single train/test split provided by the repository. We use the train split to fine-tune the parameters of the encoder (if needed), and to learn the mapping from the representations to the specific class labels. The test split is solely used to compute the reported accuracy scores. Following common practice, all time series are pre-normalized to have zero mean and unit variance.

\subsection{Measures}

In addition to the raw accuracy score (in \%), we consider the normalized accuracy ratio
\begin{equation*}
R_i = \frac{A_i-A^{\text{M}}_i}{100-A^{\text{M}}_i} ,
\end{equation*}
where $i$ denotes the $i$-th data set, $A_i$ is the accuracy obtained with the current classifier, and $A^{\text{M}}_i$ is the accuracy of a majority-based classifier. This way, $R$ is a quantity that is normalized by both the number of classes and the relative difficulty of the prediction task with respect to the class distribution. Apart from $A$ and $R$, we also report the average rank of the considered approaches, including the baselines evaluated in the repository, and the number of times an approach is the best across all approaches and baselines. We find a total of 36~baselines in the repository, including some of the most competitive existing approaches~\cite{Bagnall17DMKD}.

\subsection{Encoder Adaptation}

To assess the goodness of the learned representations in the case of no adaptation, we consider the performance of a one nearest-neighbor (1NN) classifier. The 1NN classifier is the main choice to evaluate time series similarity measures~\cite{Serra14KBS}, and almost always outperforms other classifiers when considering the raw time series~\cite{Bagnall17DMKD}. In our case, the 1NN classifier performs no further adaptation or learning (it only retrieves closest points), and exploits the Euclidean distance between representations, which we believe is an interesting proxy for other unsupervised tasks like clustering or motif discovery. 

To assess the goodness of the learned representations in the case of performing some adaptation, we consider the performance of two classifiers\footnote{Unless stated otherwise, we use scikit-learn~\cite{Pedregosa11JMLR} version 0.19.1 with default parameters.}: a logistic regression classifier (LR; with regularization or complexity parameter $C=0.1$) and a support vector machine with a radial basis function kernel (SVM; $C=100$). 
In all previous cases, the parameters of the encoder remain frozen while the classifiers learn to map representations to class labels. We only normalize the representation components to have zero mean and unit variance. 

A further case we consider is the adaptation of both encoder and mapping to the new task. For that we take the pre-trained encoder and fine-tune it, together with a fully-connected layer with softmax activation (ADAPT). Finally, to assess the benefit of encoder pre-training, we also consider an additional encoder network trained from scratch solely on the new target task (NEW). 
Training for ADAPT and NEW is done with Adam~\cite{Kingma15ICLR} for 100~epochs with empirically-chosen learning rates of $5\cdot 10^{-5}$ and $10^{-4}$, respectively. In pre-analysis, we made sure that both ADAPT and NEW were able to converge to a stable solution with this amount of training.


\section{Results}

\subsection{Accuracy and Ranking}

As mentioned, we compute the evaluation measures for every encoder-based approach on all the 85~data sets of the UCR/UEA repository, and then compare against all 36~baselines available in the same repository (Sec.~\ref{sec:eval}). However, due to space constrains, and for ease of summarization, we only report average measures and focus on selected baseline approaches (Table~\ref{tab:main_res}). First of all, we observe that using the raw learned representation without adaptation (Encoder-1NN) is already a very competitive strategy. It clearly outperforms the Euclidean distance baseline, and has a better accuracy than classical distance measures like DTW. Moreover, it obtains accuracies comparable to the top-scoring similarity measures (TWE and MSM). 

\begin{table}[t]\setlength{\tabcolsep}{10pt}
	\begin{tabular}{p{2.2cm}rrrr}
\hline\hline
Approach		& \multicolumn{1}{c}{$\bar{A}$} & \multicolumn{1}{c}{$\bar{R}$}	& Rank & Wins \\
\hline\hline
Euclidean-1NN   &  70.9 & 0.504 & 29.7 &  1 \\
DTW-Rn-1NN      &  75.9 & 0.580 & 23.4 &  2 \\
TWE-1NN         &  76.4 & 0.580 & 22.4 &  3 \\
\bf Encoder-1NN & \bf  76.5 & \bf 0.599 & \bf 22.7 & \bf  2 \\
MSM-1NN         &  77.3 & 0.593 & 20.1 &  2 \\
\hline
RotF            &  77.6 & 0.608 & 17.8 &  6 \\
\bf Encoder-LR  & \bf  79.8 & \bf 0.650 & \bf 17.3 & \bf  5 \\
\bf Encoder-SVM & \bf  80.3 & \bf 0.667 & \bf 15.6 & \bf  5 \\
\hline
BOSS            &  81.0 & 0.676 & 14.3 & 15 \\
\bf Encoder-NEW & \bf  81.3 & \bf 0.682 & \bf 11.9 & \bf 16 \\
ST              &  82.2 & 0.694 & 11.9 & 17 \\
\bf Encoder-ADAPT & \bf  82.9 & \bf 0.708 & \bf  8.7 & \bf 26 \\
COTE            &  83.8 & 0.715 &  7.7 & 18 \\
\hline\hline
	\end{tabular}
	\vspace*{0.2cm}
	\caption{Average performance of selected approaches. Values are computed by considering the original single splits of all the 85~data sets and 36~baselines of the UCR/UEA repository, together with the encoder-based approaches. However, due to space constrains, we do not show all baselines and individual data set values. The encoder-based classifiers use $k=256$.}
	\label{tab:main_res}
\end{table}

These results are interesting because we are using plain Euclidean distance over learned representations. Given that results are comparable to or better than current distance-based approaches, the advantage of using the encoder-based representation over the raw time series is essentially threefold. First, representations are generally more compact than the raw time series. Here use representations of $k=256$~numbers, which corresponds to a reduced-size representation for more than half of the training sets available in the repository. Second, representations are fast to compute, in the order of milliseconds with a Titan~Xp GPU for a hundred time series. Third, the use of Euclidean distance is quite appealing, as it is already implemented in almost all data processing libraries, with efficient methods to deal with nearest-neighbor queries.

Going back to the main results (Table~\ref{tab:main_res}), we observe that, if we learn a mapping from representations to classes while keeping the encoder weights frozen, the encoder-based architectures outperform dedicated classifiers with the raw time series as input (RotF). If we further adapt the encoder weights to a specific classification task, we observe that the resulting approach (Encoder-ADAPT) is competitive with the state-of-the-art. Only the best baselines beat the obtained classifiers (COTE and, in principle, HIVE-COTE~\cite{Lines16ICDM}, which is not available in the repository). These are ensemble-based methods that, compared to adapting the encoder architecture, might presumably be significantly less efficient, both at training and at testing time~\cite{Bagnall17DMKD}. A further interesting thing to note is that Encoder-ADAPT outperforms COTE in number of wins, but overall has a lower average rank. This indicates that Encoder-ADAPT can perform well on a number of data sets but, nonetheless, performs poorly on others. In future work, we plan to gain insight on this question. Finally, we also observe that starting from a pre-trained encoder (Encoder-ADAPT) is better than training from scratch the exactly same architecture only with the target data set (Encoder-NEW).

\subsection{Effect of Representation Size}
\label{sec:res_encoding_dim}

\begin{figure}[t]
	\includegraphics{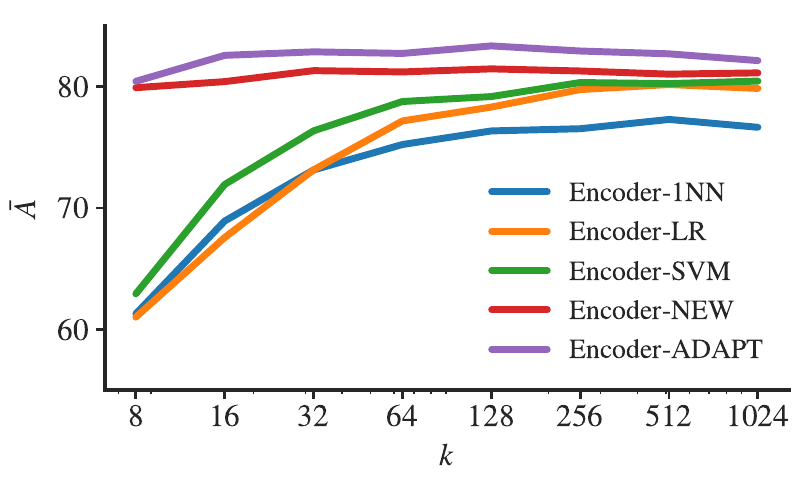}
	\vspace*{-0.3cm}
	\caption{Effect of the encoding size. Average accuracy $\bar{A}$ as a function of representation dimensionality $k$. We do not consider values of $k>1024$ as then the time series representation would be larger than almost all raw time series, thus expanding the size of the data instead of compacting it.}
	\label{fig:nfeat}
\end{figure}

We can also study how the size of the representations $k$ affects the final accuracy (Fig.~\ref{fig:nfeat}). Overall, we observe two trends, which correspond to the fact of adapting or not adapting the encoder network to the target test set. If we do not adapt the encoder network (Encoder-1NN, Encoder-LR, and Encoder-SVM), we see that, the lower the representation dimensionality, the lower the performance of the classifiers. This is to be expected, as with lower $k$ the encoder is forced to tradeoff potentially relevant information for compactness. Contrastingly, if we adapt the encoder network (Encoder-NEW and Encoder-ADAPT), we see that the representation dimensionality does not have a clear effect on the results. There seems to be a marginally optimal operation point between $k=64$ and $k=256$, but the difference with the rest of operation points might not be significant.

\subsection{Informal Report of Alternative Architectures}

To develop the proposed encoder architecture, we started from the successful convolutional network by Wang et al.~\cite{Wang16ARXIV}. However, we found that the proposed attention strategy outperformed the original global average pooling strategy, specially for Encoder-1NN. In addition, we replaced batch normalization by instance normalization, and added a final instance normalization layer. We again found the latter to substantially help in the case of Encoder-1NN, Encoder-LR, and Encoder-SVM. An additional change with respect to that work is the introduction of max-pooling, which increased the efficiency of the encoder, and the use of larger convolutional kernel sizes, which we found yield slightly better accuracies.

In addition to the aforementioned architectures, we also experimented with a number of alternative strategies. One of the non-successful strategies we tried was to substitute the attention mechanism by a recurrent neural network. With that, we could achieve marginally better accuracies in the validation set that, nonetheless, did not generalize well to the out-of-type test sets. A further non-successful architecture change we considered was the use of causal dilated convolutions~\cite{Oord16ARXIV} with padding.


\section{Conclusion and Future Work}

We have studied the use of a universal encoder for time series in the specific case of classifying an out-of-sample data set of an unseen data type. We have considered the cases of no-adaptation, mapping adaptation, and full adaptation. In all cases we achieve performances that are competitive with the state-of-the-art that, in addition, involve a compact reusable representation and few training iterations. We have also studied the effect of the representation dimensionality, showing that small representations have an impact to no-adaptation and mapping adaptation approaches, but not much to full adaptation ones.

In the future, we plan to refine the encoder architecture, as well as optimizing some of the parameters we empirically use in our experiments. A very interesting direction for future research is the adoption of one-shot learning schemas~\cite{Snell17NIPS,Vinyals16NIPS}, which we find very suitable for the current setting in time series classification problems. A further option to enhance the performance of a universal encoder is data augmentation, specially considering recent linear instance/class interpolation approaches~\cite{Zhang18ICLR}.


\bibliography{biblio}
\bibliographystyle{plainurl}

\end{document}